\documentclass[10pt,twocolumn,letterpaper]{article}

\usepackage[pagenumbers]{cvpr} 

\definecolor{cvprblue}{rgb}{0.21,0.49,0.74}
\usepackage[pagebackref,breaklinks,colorlinks,allcolors=cvprblue]{hyperref}

\usepackage{booktabs}
\usepackage{multirow}
\usepackage{algorithm}
\usepackage{algorithmic}
\usepackage{graphicx, amsmath, amssymb, caption, subcaption, multirow, overpic, textpos}
\usepackage{arydshln} 

\makeatletter
\usepackage{xspace}
\def\@onedot{\ifx\@let@token.\else.\null\fi\xspace}
\DeclareRobustCommand\onedot{\futurelet\@let@token\@onedot}

\newcommand{\tabref}[1]{Tab\onedot~\ref{#1}}

\def\eg{\emph{e.g}\onedot} 
\def\ie{\emph{i.e}\onedot} 
 
 \def\vs{\emph{vs}\onedot}
\def\wrt{w.r.t\onedot}

\newcommand{\figref}[1]{Fig\onedot~\ref{#1}}
\newcommand{\secref}[1]{Sec\onedot~\ref{#1}}
\usepackage{xcolor}
\usepackage{colortbl}
\definecolor{baselinecolor}{gray}{.9}
\newcommand{\baseline}[1]{\cellcolor{baselinecolor}{#1}}

\newlength\savewidth\newcommand\shline{\noalign{\global\savewidth\arrayrulewidth
  \global\arrayrulewidth 1pt}\hline\noalign{\global\arrayrulewidth\savewidth}}
\newcommand{\tablestyle}[2]{\setlength{\tabcolsep}{#1}\renewcommand{\arraystretch}{#2}\centering\footnotesize}

\def\paperID{****} 
\def\confName{CVPR}
\def\confYear{2025}

\title{Randomized Autoregressive Visual Generation}

\author{Qihang Yu
\;\;\;\;
Ju He
\;\;\;\;
Xueqing Deng
\;\;\;\;
Xiaohui Shen
\;\;\;\;
Liang-Chieh Chen \\
ByteDance\\
\url{https://yucornetto.github.io/projects/rar.html}
}

\begin{document}
\maketitle
\begin{abstract}
This paper presents \textbf{R}andomized \textbf{A}uto\textbf{R}egressive modeling (\textbf{RAR}) for visual generation, which sets a new state-of-the-art performance on the image generation task while maintaining full compatibility with language modeling frameworks. The proposed RAR is simple: during a standard autoregressive training process with a next-token prediction objective, the input sequence—typically ordered in raster form—is randomly permuted into different factorization orders with a probability $r$, where $r$ starts at $1$ and linearly decays to $0$ over the course of training. This annealing training strategy enables the model to learn to maximize the expected likelihood over all factorization orders and thus effectively improve the model's capability of modeling bidirectional contexts. Importantly, RAR preserves the integrity of the autoregressive modeling framework, ensuring full compatibility with language modeling while significantly improving performance in image generation. On the ImageNet-256 benchmark, RAR achieves an FID score of \textbf{1.48}, not only surpassing prior state-of-the-art autoregressive image generators but also outperforming leading diffusion-based and masked transformer-based methods. Code and models will be made available at \url{https://github.com/bytedance/1d-tokenizer}.

\end{abstract}    
\section{Introduction}
AutoRegressive (AR) models have driven remarkable advancements across both natural language processing and computer vision tasks in recent years. In language modeling, they serve as the fundamental framework for Large Language Models (LLMs) such as GPT~\cite{openai2023gpt}, Llama~\cite{touvron2023llama,touvron2023llama2}, and Gemini~\cite{team2023gemini}, along with other state-of-the-art models~\cite{yang2024qwen2,abdin2024phi}. In the realm of computer vision,  autoregressive models\footnote{While MaskGIT-style models~\cite{chang2022maskgit} could be classified as ``generalized autoregressive models'' as defined in~\cite{li2024autoregressive}, in this paper, we primarily use the term ``autoregressive'' to refer to GPT-style models~\cite{esser2021taming,yu2021vector,sun2024autoregressive}, which are characterized by \textbf{\emph{causal}} attention, \textbf{\emph{next-token}} prediction, and operate \textbf{\emph{without}} the need for mask tokens as placeholders.}
have also shown substantial potential, delivering competitive performance in image generation tasks~\cite{esser2021taming,yu2021vector,yu2022scaling,lee2022autoregressive,sun2024autoregressive,luo2024open} to diffusion models~\cite{dhariwal2021diffusion,bao2023all,peebles2023scalable,li2024autoregressive} or non-autoregressive transformers~\cite{chang2022maskgit,yu2023magvit,yu2023language,yu2024image,weber2024maskbit}. More importantly, autoregressive modeling is emerging as a promising pathway toward unified models across multiple modalities and tasks~\cite{brown2020language,wei2021finetuned,chung2024scaling,bai2023sequential,team2024chameleon,team2024emu3}.

\begin{figure}[t!]
    \centering
    \includegraphics[width=1.0\linewidth]{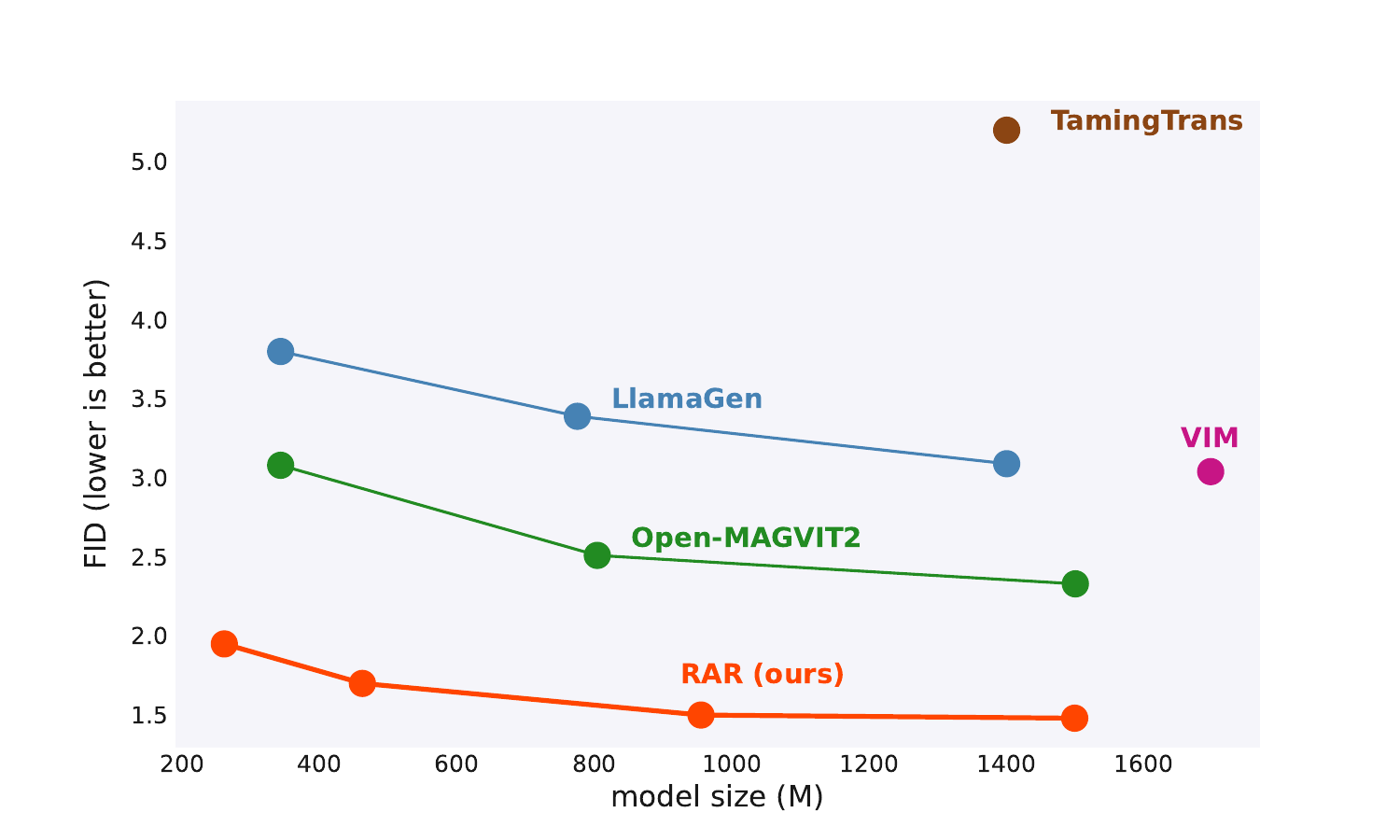}
    \caption{
    \textbf{Comparison among different language modeling compatible autoregressive (AR) image generators.}
    The proposed RAR demonstrates significant  improvements over previous AR methods. RAR-B, with only \textit{261M} parameters, achieves an FID score of 1.95, outperforming both LlamaGen-XXL (1.4B parameters) and Open-MAGVIT2-XL (1.5B parameters). 
    }
    \label{fig:ar_compare}
\end{figure}
\label{sec:intro}

\begin{figure*}[t!]
    \centering
    \includegraphics[width=1.0\linewidth]{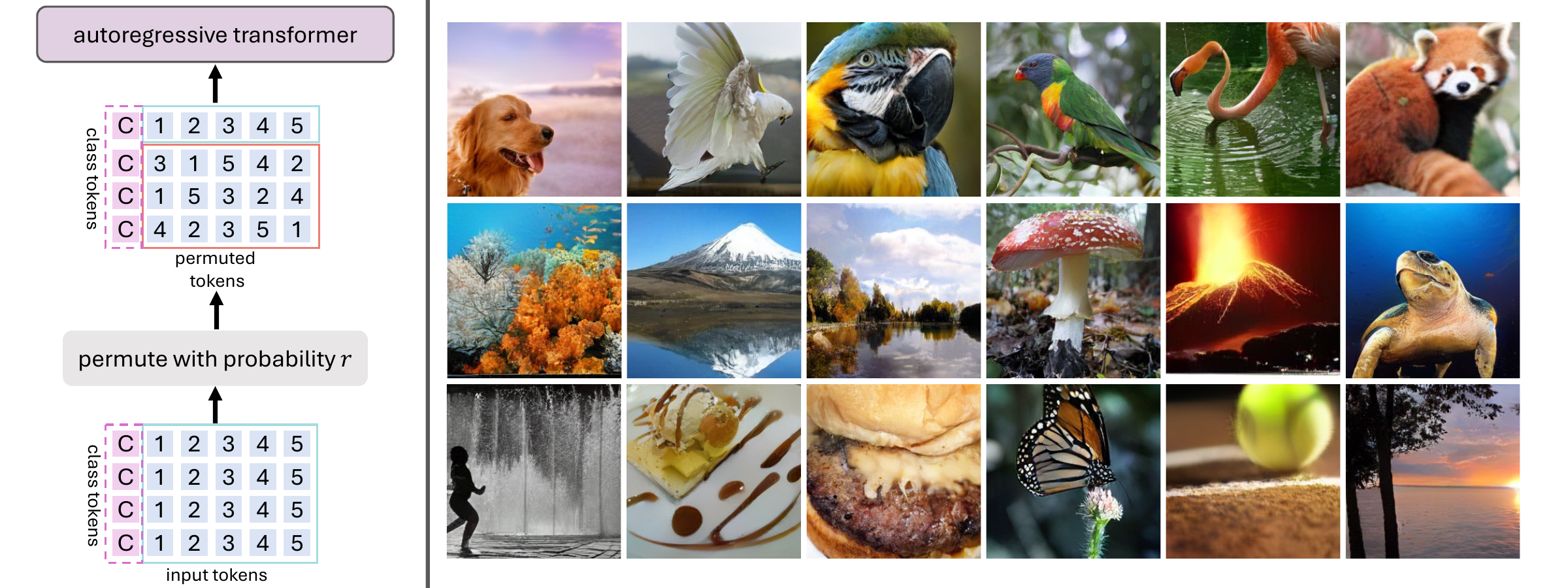}
    \caption{
    \textbf{Overview of the proposed Randomized AutoRegressive (RAR) model, which is fully compatible with language modeling frameworks.}
    \textit{Left}: RAR introduces a randomness annealing training strategy to enhance the model's ability to learn bidirectional contexts. During training, the input sequence is randomly permuted with a probability $r$, which starts at 1 (fully random permutations) and linearly decreases to 0, transitioning the model to a fixed scan order, such as raster scan, by the end of training.
    \textit{Right}: Randomly selected images generated by RAR, trained on ImageNet.
    }
    \label{fig:framework}
\end{figure*}

Despite the dominance of autoregressive models in language modeling, they often yield suboptimal performance in comparison to diffusion models or non-autoregressive transformers in visual generation tasks~\cite{sun2024autoregressive,luo2024open}. This discrepancy can be attributed to the inherent differences between text and visual signals. Text is highly compact and semantically meaningful, while visual data tends to be more low-level and redundant~\cite{he2022masked,yu2024image}, making bidirectional context modeling more critical. For instance, several studies~\cite{li2024autoregressive,el2024scalable,beyer2024paligemma} have demonstrated that causal attention applied to image tokens leads to inferior performance compared to bidirectional attention in vision tasks.

To address this, recent works~\cite{tian2024visual,li2024autoregressive} have attempted to reintroduce bidirectional attention by redesigning the autoregressive formulation, achieving state-of-the-art results in image generation. However, these approaches often deviate from the traditional autoregressive paradigm. For example, VAR~\cite{tian2024visual} shifts from next-token prediction to next-scale prediction, enabling bidirectional attention within each scale, and MAR~\cite{li2024autoregressive} generalizes MaskGIT-style framework~\cite{chang2022maskgit} to the autoregressive definition, which naturally introduces back the bidirectional attention. While effective, these modifications complicate their integration into universal transformer architectures that aim to unify different modalities, which proves to work well with conventional autoregressive modeling~\cite{team2024chameleon,team2024emu3}.

In this paper, we aim to enhance the generation quality of autoregressive image models while preserving the core autoregressive structure, maintaining compatibility with language modeling frameworks. Specifically, we enable bidirectional context learning within an autoregressive transformer by maximizing the expected likelihood over all possible factorization order. In this way, all tokens will be trained and predicted under all possible contexts,  facilitating learning bidirectional representation.
Moreover, we introduce a permutation probability $r$, which controls the ratio of training data between a random factorization order and the standard raster order. Initially, $r$ is set to $1$ (fully random factorization) and it linearly decays to $0$ over the course of training, gradually reverting the model to the raster order commonly used by other autoregressive image generators.

To this end, we present a simple, effective, and scalable autoregressive model training paradiam named \textbf{R}andomized \textbf{A}uto\textbf{R}egressive modeling (\textbf{RAR}). RAR retains the original autoregressive model architecture and formulation, ensuring full compatibility with language modeling. At the same time, it significantly improves the generation quality of autoregressive models at no additional cost.
On the ImageNet-256 benchmark~\cite{deng2009imagenet}, RAR achieves an FID score of $1.48$, substantially outperforming previous state-of-the-art autoregressive image generators, as illustrated in~\figref{fig:ar_compare}. By addressing the limitations of unidirectional context modeling, RAR represents a critical step towards autoregressive visual generation and opens up new possibilities for further advancements in the field.

\section{Related Work}
\label{sec:related}

\noindent \textbf{Autoregressive Language Modeling.} 
The advent of autoregressive language models~\cite{radford2018improving,radford2021learning,brown2020language,openai2023gpt,achiam2023gpt,touvron2023llama,touvron2023llama2,dubey2024llama,team2023gemini,chowdhery2023palm,anil2023palm,bai2023qwen,yang2024qwen2,abdin2024phi} has paved a promising path toward general-purpose AI systems. At the core of these models is a simple yet powerful next-token prediction paradigm, where the objective is to predict the next word or token in a sequence based on preceding inputs. This approach has demonstrated both scalability, as evidenced by scaling laws, and versatility through zero-shot generalization, enabling explorations beyond traditional language tasks to diverse modalities.

\vspace{0.5ex}
\noindent \textbf{Autoregressive Visual Modeling.}
Pioneering research~\cite{gregor2014deep,van2016pixel,van2016conditional,parmar2018image,chen2020generative} in autoregressive visual modeling has focused on representing images as sequences of pixels. Nevertheless, inspired by advancements in autoregressive language modeling, a subsequent wave of studies has transitioned to modeling images as sequences of discrete-valued tokens~\cite{van2017neural,razavi2019generating,esser2021taming,ramesh2021zero,yu2021vector}, resulting in notable improvements in performance. This direction has been further explored through efforts~\cite{sun2024autoregressive,luo2024open} aimed at enhancing tokenization quality and leveraging modern autoregressive architectures initially developed for language tasks.
However, all of these works strictly adhere to a raster-scan order for processing pixels or tokens, resulting in a unidirectional information flow that is sub-optimal for visual modeling. In this work, we instead explore learning across all possible factorization orders to enhance bidirectional context learning while retaining the core autoregressive framework.

\vspace{0.5ex}
\noindent \textbf{Other Visual Generation Models.} 
In addition to autoregressive visual modeling, there have been numerous efforts in exploring other formats of visual generation models, including generative adversarial networks (GANs)~\cite{goodfellow2014generative, brock2018large,karras2019style}, diffusion models~\cite{ho2020denoising,dhariwal2021diffusion,rombach2022high,peebles2023scalable,liu2024alleviating,esser2024scaling}, masked transformers~\cite{chang2022maskgit,chang2023muse,yu2023magvit,yu2024image,weber2024maskbit}, scale-wise autoregressive modeling (VAR)~\cite{tian2024visual,ma2024star,zhang2024var,tang2024hart}, and masked autoregressive modeling with diffusion loss (MAR)~\cite{li2024autoregressive,fan2024fluid}. It is worth noting that MAR~\cite{li2024autoregressive} also experimented a random order based AR framework similar to the proposed RAR. However, as indicated in our experiments (see~\secref{sec:ablation_studies}), simply replacing the raster order with random order only brings marginal improvement, coinciding the observation in~\cite{li2024autoregressive}. This further demonstrates the importance on the randomness annealing strategy in RAR, leading to a substantial improvement for the AR image generators.

\section{Method}
\label{sec:method}
In this section, we first provide an overview of autoregressive modeling in~\secref{sec:background}, followed by our proposed Randomized AutoRegressive modeling (RAR) in~\secref{sec:rar}.

\begin{figure*}[t!]
    \centering
    \vspace{-15ex}
    \includegraphics[width=1.0\linewidth]{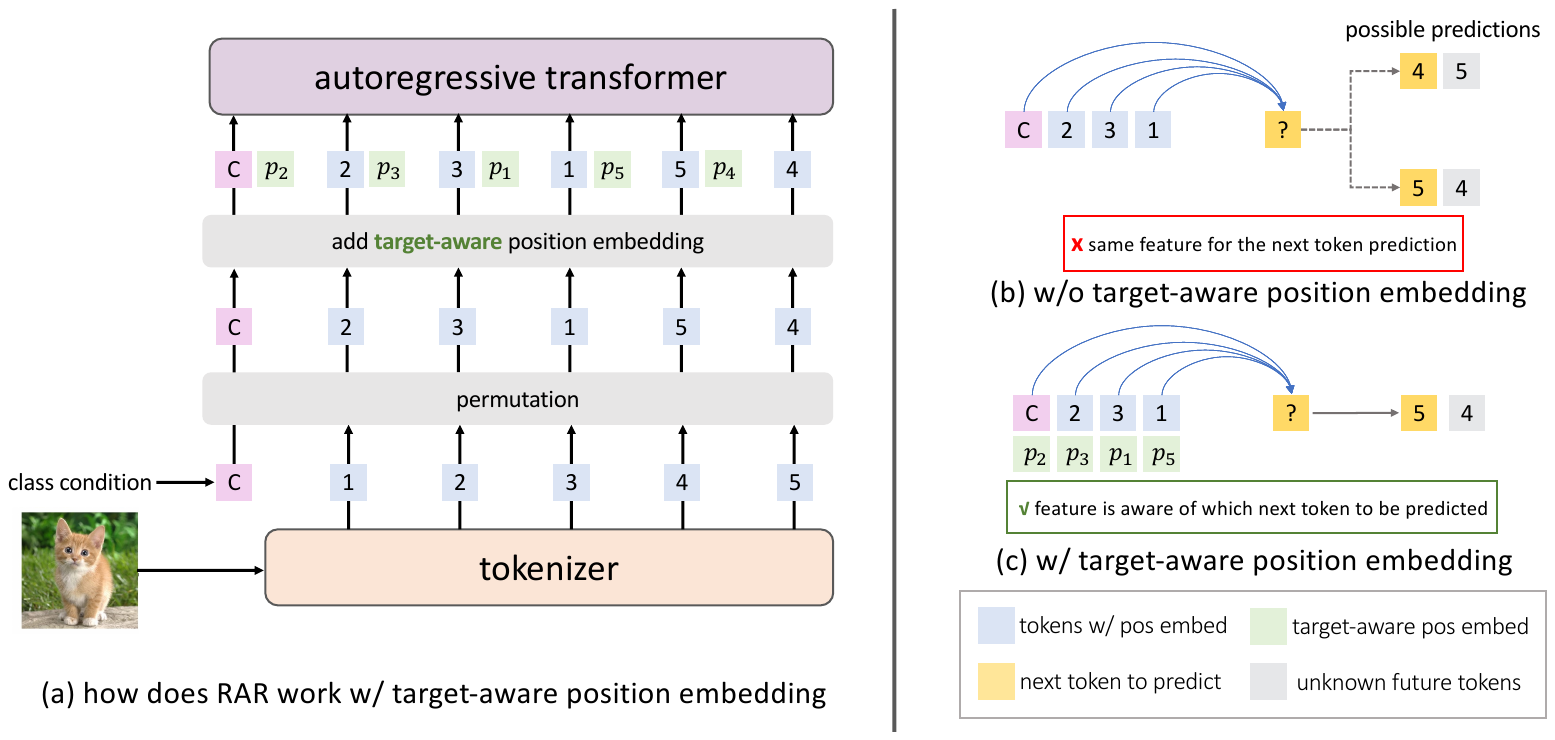}
    \vspace{-20ex}
    \caption{
    \textbf{Illustration of the target-aware positional embedding.} Subfigure (a) shows the training process of the proposed Randomized AutoRegressive (RAR) model, along with the target-aware position embedding.
    Following Vision Transformer~\cite{dosovitskiy2020image}, images are tokenized into patches with  original position embeddings (blue tokens). 
    The token sequence is then randomly permuted, with the target-aware positional embeddings (green tokens) added to guide the model. 
    Subfigures (b) and (c) highlight the importance of the target-aware positional embedding: (b) demonstrates a failure case where both permuted sequences yield identical prediction logits, while (c) shows that the target-aware positional embedding correctly guides the model to predict the next token accurately.
    }
    \label{fig:target_aware}
\end{figure*}

\subsection{Background}
\label{sec:background}
We provide a brief overview of autoregressive modeling with a next-token prediction objective. Given a discrete token sequence $\mathbf{x}=[x_1, x_2, \cdots, x_T]$, the goal of autoregressive modeling is to maximize the likelihood of the sequence under a forward autoregressive factorization. Specifically, the objective is to maximize the joint probability of predicting the current token $x_t$ based on all preceding tokens $[x_1, x_2, \cdots, x_{t-1}]$, $\forall t=1, \cdots, T$:
\begin{equation}
\underset{\theta}{\mathrm{max}}\quad p_{\theta}(\mathbf{x}) = \prod_{t=1}^Tp_{\theta}(x_t | x_1, x_2, \cdots, x_{t-1}),
\label{old_ar_obj}
\end{equation}
where $p_{\theta}$ denotes a token distribution predictor with a model parameterized by $\theta$.

As shown in the equation, each token $x_t$ at position $t$ is conditioned solely on the preceding tokens, which limits context modeling to a unidirectional manner. This contrasts with methods such as masked transformer~\cite{chang2022maskgit,yu2023magvit,yu2023language,weber2024maskbit} and diffusion models~\cite{ho2020denoising,rombach2022high,peebles2023scalable,liu2024alleviating}, which can leverage bidirectional context at the training time. Additionally, while natural language has an inherent sequential order (left-to-right in most languages), image data lacks a clear, predefined order for processing tokens. Among the possible orders for image generation, the row-major order (\ie, raster scan) is the most widely adopted and has demonstrated superior performance compared to other alternatives~\cite{esser2021taming}.

\subsection{RAR: Randomized AutoRegressive Modeling}
\label{sec:rar}
Visual signals inherently exhibit bidirectional correlations, making effective global context modeling essential. However, conventional autoregressive models rely on causal attention masking, which enforces a unidirectional dependency on the token sequence, contradicting the nature of visual data, as noted in prior works~\cite{beyer2024paligemma,el2024scalable,li2024autoregressive}, where bidirectional attention works significantly better than causal attention for visual modality. Furthermore, there is no universally ``correct" way to arrange image tokens into a causal sequence. While the widely adopted raster order has achieved some success, it introduces biases in the autoregressive training process. For instance, each token is conditioned solely on the preceding tokens in the scanning order, restricting the model's ability to learn dependencies from other directions.

To address these challenges, we propose a randomized autoregressive modeling approach that incorporates optimization objective with bidirectional context:
\begin{equation}
\underset{\theta}{\mathrm{max}}\quad p_{\theta}(\mathbf{x}) = \prod_{t=1}^Tp_{\theta}(x_t | x_1, \cdots, x_{t-1}, x_{t+1}, \cdots, x_T).
\label{new_ar_obj}
\end{equation}

Unlike BERT-style~\cite{devlin2018bert} or MaskGIT-style~\cite{chang2022maskgit} methods, our method follows the permuted objective approach~\cite{uria2016neural,yang2019xlnet}, where the model is trained in an autoregressive manner across all possible factorization orders. This enables the model to gather bidirectional context while preserving the autoregressive framework {\it in expectation}. Formally, we have:
\begin{equation}
\underset{\theta}{\mathrm{max}}\quad p_{\theta}(\mathbf{x}) = \mathbb{E}_{\tau \sim \mathcal{S}_T}\left[ \prod_{t=1}^Tp_{\theta}(x_{\tau_t} | x_{\tau_{<t}})\right],
\label{permute_training}
\end{equation}
where $\mathcal{S}_T$ denotes the set of all possible permutations of the index sequence $[1, 2, \cdots, T]$, and $\tau$ represents a randomly sampled permutation from $\mathcal{S}_T$. The notation $\tau_t$ refers to the $t$-th element in the permuted sequence, and $\tau_{<t}$ represents all preceding positions to $\tau_t$. Since the model parameters $\theta$ are shared across all sampled factorization orders, each token $x_t$ is exposed to every possible context and learns relationships with every other token $x_i$ $\forall i \neq t$,  during training. This allows the model to effectively capture bidirectional context while preserving the integrity of the autoregressive formulation.

Although simple, this modification significantly improves image generation performance, highlighting the power of bidirectional context in improving autoregressive image generator capability. Our findings align with those observed in autoregressive training for language modeling in NLP~\cite{uria2016neural,devlin2018bert,yang2019xlnet,brown2020language} as well.

\vspace{1ex}
\noindent \textbf{Discussion.} While the permutation objective allows for bidirectional context learning within the autoregressive framework {\it in expectation}, it remains challenging to fully capture ``global context" during the generation process. This is because there are always some tokens generated before others, without having access to the full global context. This limitation is not unique to autoregressive methods~\cite{esser2021taming,sun2024autoregressive} but also present in non-autoregressive models~\cite{chang2022maskgit}. Techniques such as resampling or refinement~\cite{gu2019levenshtein,madaan2024self} may help address this issue by ensuring that every token is generated with sufficient context. However, such designs may complicate the system; thus, exploring such solutions lies beyond the scope of this paper and is left for future work.

\vspace{1ex}
\noindent \textbf{Target-aware Positional Embedding.} One limitation of the permuted training objective is that standard positional embeddings may fail in certain scenarios. For instance, consider two different permutations: $\mathbf{\tau}_a = [1, 2, \cdots, T-2, T-1, T]$ and $\mathbf{\tau}_b = [1, 2, \cdots, T-2, T, T-1]$ (\ie, only the last two tokens' positions are swapped). When predicting the second to last token, both permutations will yield identical features and thus identical prediction logits, even though they correspond to different ground-truth labels (\ie, $p_{\theta}(x_{\tau_{T-1}} | x_{\tau_1}, x_{\tau_2}, \cdots, x_{\tau_{T-2}})$ is the same for both permutations $\mathbf{\tau}_a$ and $\mathbf{\tau}_b$). This problem, in a general randomized autoregressive training process and beyond this specific example, can happen for all token locations except the last one (since the last token does not need to predict next token). To address this issue, we introduce an additional set of positional embeddings, which we refer to as {\it target-aware positional embeddings}. These embeddings encode information about which token is being predicted next.

Formally, we define a set of target-aware positional embeddings $\mathbf{p}_{ta} = [p_1, p_2, \cdots, p_T]$. The positional embedding corresponding to the next token is added to the current token embedding, resulting in a target-aware token embedding $\hat{\mathbf{x}}_\tau$:
\begin{equation}
\hat{\mathbf{x}}_{\tau} = \mathbf{x}_{\tau} + \mathbf{p}_{\tau} = [x_{\tau_1} + p_{\tau_2}, x_{\tau_2} + p_{\tau_3}, \cdots, x_{\tau_{T-1}} + p_{\tau_T}, x_{\tau_T}],
\label{eq:target_aware}
\end{equation}
where $\mathbf{x}_{\tau}$ and $\mathbf{p}_{\tau}$ are permuted tokens for $\mathbf{x}$ and $\mathbf{p}_{ta}$ \wrt to the permutation $\tau$, respectively. By associating the target token's positional embedding with the next-token prediction, each token prediction is aware of the target token's index, alleviating the potential confusion in permuted objective.

Notably, we omit the target-aware positional embedding for the final token $x_{\tau_T}$, as it does not participate in the loss computation and has no prediction target. A visual illustration of this concept is provided in~\figref{fig:target_aware}.
It is also noteworthy that the target-aware positional embedding can be merged with original positional embedding after the training is finished, because our method anneals to a fixed raster scan in the end, and thus leads to no increase on the parameters or computation during inference.

\vspace{1ex}
\noindent \textbf{Randomness Annealing.} While the proposed randomized autoregressive training with permutation enables the model to capture bidirectional context within a unidirectional framework, it may introduce sub-optimal behavior for visual generation due to two main factors: (1) The sheer number of possible permutations is vast, potentially causing the model to focus on learning how to handle the different permutation orders rather than improving generation quality. For example, for a token sequence of length $256$, the number of possible permutations is $256! > 10^{506}$, which can overwhelm the model and reduce training efficiency. (2) Although images can be processed in arbitrary orders, certain scan orders tend to outperform others. For instance, \cite{esser2021taming} evaluated six different scan orders (row-major, spiral in, spiral out, z-curve, subsample, and alternate) and found that row-major (\ie, raster order) consistently performed the best, a result that has made it the most widely used order for visual generation.

To address these issues, we propose Randomness Annealing, a strategy designed to balance the randomness of permutations with the known effectiveness of the raster order. This method introduces a single parameter, $r$, which controls the probability of using a random permutation versus the raster order. At the start of training, $r = 1$, meaning that the model exclusively uses random permutations. Over the course of training, $r$ linearly decays to $0$, transitioning the model to the raster order by the end of training. Specifically, we define a training schedule for $r$, controlled by two hyper-parameters $start$ and $end$ indicating the training epoch when $r$ starts to anneal and when the annealing ends. Formally, we have:
\begin{equation}
r = \begin{cases}
    1.0, & \text{if } epoch < start,\\
    0.0, & \text{if } epoch > end,\\
    1.0 - \frac{epoch - start}{end - start}, & \text{otherwise},
\end{cases}
\label{eq:r_schedule}
\end{equation}
where $epoch$ is the current training epoch. We will ablate the hyper-parameters $start$ and $end$ in the experiments.

The schedule allows the model to initially explore the diverse random permutations for better bidirectional representation learning, and ultimately converge to the more effective row-major scan order for better visual generation quality, as is used by other typical autoregressive methods~\cite{esser2021taming}. It is worth noting that this strategy not only improves generation performance but also maintains compatibility with the standard scan order used in previous works.
\section{Experimental Results}
\label{sec:experiment}
In this section, we  outline the implementation details of our method in~\secref{sec:implementation_details}. Next, we present ablation studies on key design choices in~\secref{sec:ablation_studies}. The main results are discussed in~\secref{sec:main_results}, followed by scaling study and visualizations.

\subsection{Implementation Details}
\label{sec:implementation_details}
We implement the RAR on top of language modeling autoregressive framework with minimal changes.

\vspace{0.5ex}
\noindent \textbf{VQ Tokenizer.} Following prior works~\cite{esser2021taming,chang2022maskgit} which use a VQ tokenizer to tokenize the input images into discrete token sequences, we use the MaskGIT-VQGAN~\cite{chang2022maskgit} with the official weight trained on ImageNet. This tokenizer is a purely CNN-based tokenizer which tokenizes a $256\times 256$ image into $256$ discrete tokens (\ie, downsampling factor 16) with a codebook size (\ie, vocabulary size) 1024. 

\begin{table}
\centering
\tablestyle{5.0pt}{1.05}

\begin{tabular}{l|ccccc}
model & depth & width & mlp & heads & \#params  \\
\shline
RAR-B & 24 & 768 & 3072 & 16  & 261M \\
RAR-L & 24 & 1024 & 4096 & 16  & 461M  \\
RAR-XL & 32 & 1280 & 5120 & 16  & 955M  \\
RAR-XXL & 40 & 1408 & 6144 & 16  & 1499M  \\

\end{tabular}
\caption{\textbf{Architecture configurations of RAR.} We follow prior works scaling up ViT~\cite{dosovitskiy2020image,zhai2022scaling} for different configurations.
}
\label{tab:arch}
\end{table}

\vspace{0.5ex}
\noindent \textbf{Autoregressive Transformer.} We use vision transformers~\cite{dosovitskiy2020image} of different model configurations~\cite{zhai2022scaling} including RAR-S (133M), RAR-B (261M), RAR-L (461M), RAR-XL (955M), and RAR-XXL (1499M). For all of these model variants, we apply causal attention masking in the self-attention module and QK LayerNorm~\cite{dehghani2023scaling} to stabilize the large-scale model training. We use plain ViT for all ablation studies to speed up the experiments, and we enhance the model with adaLN~\cite{peebles2023scalable} for final models. The detailed architecture configuration and model size are available at~\tabref{tab:arch}.

\vspace{0.5ex}
\noindent \textbf{Positional Embedding.} We use learnable embeddings for both original positional embedding in ViT and target-aware positional embedding. Notably, as our model anneals to raster order-based autoregressive image generation after the training is finished, the two positional embeddings can be combined into one, making it identical to a conventional autoregressive image generator.

\vspace{0.5ex}
\noindent \textbf{Dataset.} We train our model on ImageNet-1K~\cite{deng2009imagenet} training set, which contains $1,281,167$ training images across $1000$ object classes. We pre-tokenize the whole training set with MaskGIT-VQGAN tokenizer~\cite{chang2022maskgit} to speed up the training. For ablation studies, we pre-tokenize the dataset with only center crop and horizontal flipping augmentation, while we further enhance the diversity in pretokenized datasets with ten-crop transformation~\cite{szegedy2015going,sun2024autoregressive} for final models.

\vspace{0.5ex}
\noindent \textbf{Training Protocols.} We use the same training hyper-parameters for all model variants. The model is trained with batch size $2048$ for $400$ epochs ($250k$ steps). The learning rate will be linearly increased from $0$ to $4\times 10^{-4}$ at the first $100$ epochs (warm-up), then it will be gradually decayed to $1\times 10^{-5}$ following a cosine decay schedule. We use AdamW~\cite{kingma2014adam,loshchilov2017decoupled} optimizer with beta1 $0.9$, beta2 $0.96$, and weight decay $0.03$. We perform gradient clipping with maximum gradient norm $1.0$. During training, the class condition will be dropped at a probability $0.1$. The training setting remain the same for both ablation studies and main results across all RAR model variants.

\vspace{0.5ex}
\noindent \textbf{Sampling Protocols.} We sample $50000$ images for FID computation using the evaluation code from~\cite{dhariwal2021diffusion}. We do not use any top-k or top-p based filtering techniques. We also follow prior arts~\cite{chang2023muse,gao2023masked,yu2024image} to use classifier-free guidance~\cite{ho2022classifier}. In ablation study, we use a simpler linear guidance schedule~\cite{chang2023muse} and for final models we use the improved power-cosine guidance schedule~\cite{gao2023masked}. The final detailed hyper-parameters for each model variant can be found in appendix. 

\begin{table}
\centering

\tablestyle{5.0pt}{1.05}
\begin{tabular}{cc|cccc}
start epoch & end epoch & FID$\downarrow$  & IS$\uparrow$ & Pre.$\uparrow$ & Rec.$\uparrow$ \\
\shline
0 & 0\dag & 3.08 & 245.3 & 0.85 & 0.52 \\
0 & 100 & 2.68 & 237.3 & 0.84 & 0.54 \\
0 & 200 & 2.41 & 251.5 & 0.84 & 0.54 \\
0 & 300 & 2.40 & 258.4 & 0.84 & 0.54 \\
0 & 400 & 2.43 & 265.3 & 0.84 & 0.53 \\
100 & 100 & 2.48 & 247.5 & 0.84 & 0.54 \\
100 & 200 & 2.28 & 253.1 & 0.83 & 0.55 \\
100 & 300 & 2.33 & 258.4 & 0.83 & 0.54 \\
100 & 400 & 2.39 & 266.5 & 0.84 & 0.54 \\
200 & 200 & 2.39 & 259.7 & 0.84 & 0.54 \\
\baseline{200} & \baseline{300} & \baseline{2.18} & \baseline{269.7} & \baseline{0.83} & \baseline{0.55} \\
200 & 400 & 2.55 & 241.6 & 0.84 & 0.54 \\
300 & 300 & 2.41 & 269.1 & 0.84 & 0.53 \\
300 & 400 & 2.74 & 236.4 & 0.83 & 0.54 \\
400 & 400\ddag & 3.01 & 305.6 & 0.84 & 0.52 \\

\end{tabular}
\caption{\textbf{Different \textit{start} and \textit{end} epochs for randomness annealing, with a total of 400 training epochs and model size RAR-L.} The final setting is labeled in \textcolor{gray}{gray}. \dag: When \textit{start} epoch and \textit{end} epoch are both $0$ (1st row), the training reverts to a standard raster order training. \ddag: When \textit{start} epoch and \textit{end} epoch are both $400$ (last row), the training becomes a purely random order training. After training is finished, all results are obtained with raster order sampling, except for the purely random order training (\ie, last row), where we also randomly sample the scan order following~\cite{li2024autoregressive}, which otherwise could not produce a reasonable result.
}
\label{tab:random_annealing}
\end{table}

\subsection{Ablation Studies}
\label{sec:ablation_studies}
We study different configurations for RAR, including the randomness annealing strategy and scan orders that RAR converges to.

\noindent \textbf{Randomness Annealing Strategy.}
In~\tabref{tab:random_annealing} we compare different randomness annealing strategies. 
We adopt a linear decaying schedule and focus on when should the randomization annealing {\it starts} and {\it ends} by changing two hyper-parameters {\it start} and {\it end}, as defined in Eq.~\eqref{eq:r_schedule}.
For a training lasting for $400$ epochs, we enumerate all possible combinations for every $100$ epochs.
For example, when $start = 200$ and $end = 300$, the model is trained with random permutations from $0$ to $200$ epochs and raster order from $300$ to $400$ epochs. During $200$ to $300$ epoch, the model is trained via random permutation with probability $r$ and raster order with probability $1-r$, where $r$ is computed as in Eq.~\eqref{eq:r_schedule}.
It is noteworthy that when $start = end = 0$, the model is trained with purely raster order, \ie, the standard autoregressive training.
When $start=end=400$, the model is always trained with randomly permuted input sequence.
Both cases are important baselines of the proposed randomness annealing, and they achieve FID scores of $3.08$ and $3.01$, respectively.
Interestingly, we observe all other variants achieve substantial improvement over these two baselines. For example, even simply replacing the first $100$ epochs of raster order with random permutation, it (\ie, $start =100$ and $end=100$) improves the FID to $2.48$ by $0.6$. Besides, we also note that the model prefers to keep some beginning epochs for pure random permutation training and some last epochs for better adapting to raster scan order, which usually leads to a better performance compared to other variants. All the results demonstrate that adding randomized autoregressive training with a permuted objective is beneficial to the autoregressive visual generator and leads to a boosted FID score, thanks to the improved bidirectional representation learning process.

Additionally, among all variants, we found that the case, where $start = 200$ and $end = 300$, works the best, which improves the baseline (purely raster order) FID from $3.08$ to $2.18$. This strategy allocates slightly more computes on the training with random permutation order, and focuses on the purely raster order for the last 100 epochs. Therefore, we default to adopt this annealing strategy for all RAR models.

\begin{table}
\centering
\tablestyle{8.0pt}{1.05}
\begin{tabular}{c|cccc}
scan order & FID$\downarrow$  & IS$\uparrow$ & Precision$\uparrow$ & Recall$\uparrow$ \\
\shline
\baseline{row-major} & \baseline{2.18} & \baseline{269.7} & \baseline{0.83} & \baseline{0.55} \\
spiral in & 2.50 & 256.1 & 0.84 & 0.54 \\
spiral out & 2.46 & 256.6 & 0.84 & 0.54 \\
z-curve & 2.29 & 262.7 & 0.83 & 0.55 \\
subsample & 2.39 & 258.0 & 0.84 & 0.54 \\
alternate & 2.48 & 270.9 & 0.84 & 0.53 \\
\end{tabular}
\caption{\textbf{Effect of different scan orders RAR-L converges to.} We mainly consider 6 different scan orders (row major, spiral in, spiral out, z-curve, subsample, alternate) as studied in~\cite{esser2021taming}. Our default setting is marked in \baseline{gray}. A visual illustration of different scan orders are available in the appendix.}
\label{tab:scan_orders}
\end{table}

\vspace{0.5ex}
\noindent \textbf{Different Scan Orders Besides Raster.} 
Although row-major order (\ie, raster scan) has been the de facto scan order in the visual generation, there lacks a systematic study on how good it is compared to other scan orders. We note that the work~\cite{esser2021taming} conducted a similar study 4 years ago. However, it is worth re-examining the conclusion considering the significant progress generative models have achieved in recent years. Specifically, we consider $6$ different scan orders (row-major, spiral in, spiral out, z-curve, subsample, and alternative) following~\cite{esser2021taming} that RAR may converge to. Instead of reporting the training loss and validation loss as the comparison metric~\cite{esser2021taming}, we directly evaluate their generation performance. The results are summarized in~\tabref{tab:scan_orders}. Interestingly, we observe that all variants achieve a reasonably good score, which indicates that RAR is capable of handling different scan orders. Considering that the row-major (raster scan) still demonstrates advantages over the other scan orders, we thus use the raster scan order for all final RAR models.

\begin{table}
\centering
\tablestyle{1.0pt}{1.05}
\begin{tabular}{l|ccccccc}
tokenizer & type & generator & \#params & FID$\downarrow$  & IS$\uparrow$ & Pre.$\uparrow$ & Rec.$\uparrow$ \\
\shline

VQ~\cite{rombach2022high} & Diff. & {\scriptsize LDM-8~\cite{rombach2022high}} & 258M & 7.76 & 209.5 & 0.84 & 0.35 \\
VAE~\cite{rombach2022high} & Diff. & {\scriptsize LDM-4~\cite{rombach2022high}} & 400M & 3.60 & 247.7 & 0.87 & 0.48 \\
\cdashline{1-1}
\multirow{7}{*}{VAE~\cite{sdvae}} & \multirow{7}{*}{Diff.} & {\scriptsize UViT-L/2~\cite{bao2023all}} & 287M & 3.40 & 219.9 & 0.83 & 0.52 \\
&  & {\scriptsize UViT-H/2~\cite{bao2023all}} & 501M & 2.29 & 263.9 & 0.82 & 0.57 \\
&  & {\scriptsize DiT-L/2~\cite{peebles2023scalable}} & 458M & 5.02 & 167.2 & 0.75 & 0.57 \\
&  & {\scriptsize DiT-XL/2~\cite{peebles2023scalable}} & 675M & 2.27 & 278.2 & 0.83 & 0.57 \\
&  & {\scriptsize SiT-XL~\cite{ma2024sit}} & 675M & 2.06 & 270.3 & 0.82 & 0.59 \\
&  & {\scriptsize DiMR-XL/2R~\cite{liu2024alleviating}} & 505M & 1.70 & 289.0 & 0.79 & 0.63 \\
&  & {\scriptsize MDTv2-XL/2~\cite{gao2023masked}} & 676M & 1.58 & 314.7 & 0.79 & 0.65 \\
\hline
VQ~\cite{chang2022maskgit} & Mask. & {\scriptsize MaskGIT~\cite{chang2022maskgit}} & 177M & 6.18 & 182.1 & - & - \\
VQ~\cite{yu2024image} & Mask. & {\scriptsize TiTok-S-128~\cite{yu2024image}} & 287M & 1.97 & 281.8 & - & - \\
VQ~\cite{yu2023language} & Mask. & {\scriptsize MAGVIT-v2~\cite{yu2023language}} & 307M & 1.78 & 319.4 & - & - \\ 
VQ~\cite{weber2024maskbit} & Mask. & {\scriptsize MaskBit~\cite{weber2024maskbit}} & 305M & 1.52 & 328.6 & - & - \\
\hline
\multirow{3}{*}{VAE~\cite{li2024autoregressive}} & \multirow{3}{*}{MAR} & {\scriptsize MAR-B~\cite{li2024autoregressive}} & 208M & 2.31 & 281.7 & 0.82 & 0.57 \\
&  & {\scriptsize MAR-L~\cite{li2024autoregressive}} & 479M & 1.78 & 296.0 & 0.81 & 0.60 \\
&  & {\scriptsize MAR-H~\cite{li2024autoregressive}} & 943M & 1.55 & 303.7 & 0.81 & 0.62 \\
\hline
\multirow{2}{*}{VQ~\cite{tian2024visual}} & \multirow{2}{*}{VAR} & {\scriptsize VAR-d30~\cite{tian2024visual}} & 2.0B & 1.92 & 323.1 & 0.82 & 0.59 \\
& & {\scriptsize VAR-d30-re~\cite{tian2024visual}} & 2.0B & 1.73 & 350.2 & 0.82 & 0.60 \\
\hline
\multirow{2}{*}{VQ~\cite{esser2021taming}} & \multirow{2}{*}{AR} & {\scriptsize GPT2~\cite{esser2021taming}} & 1.4B & 15.78 & 74.3 & - & - \\
& & {\scriptsize GPT2-re~\cite{esser2021taming}} & 1.4B & 5.20 & 280.3 & - & - \\

\cdashline{1-1}
\multirow{2}{*}{VQ~\cite{yu2021vector}} & \multirow{2}{*}{AR} & {\scriptsize VIM-L~\cite{yu2021vector}} & 1.7B & 4.17 & 175.1 & - & - \\
& & {\scriptsize VIM-L-re~\cite{yu2021vector}} & 1.7B & 3.04 & 227.4 & - & - \\

\cdashline{1-1}

\multirow{3}{*}{VQ~\cite{luo2024open}} & \multirow{3}{*}{AR} & {\scriptsize Open-MAGVIT2-B~\cite{luo2024open}} & 343M & 3.08 & 258.3 & 0.85 & 0.51 \\
&  & {\scriptsize Open-MAGVIT2-L~\cite{luo2024open}} & 804M & 2.51 & 271.7 & 0.84 & 0.54 \\
&  & {\scriptsize Open-MAGVIT2-XL~\cite{luo2024open}} & 1.5B & 2.33 & 271.8 & 0.84 & 0.54 \\

\cdashline{1-1}

\multirow{8}{*}{VQ~\cite{sun2024autoregressive}} & \multirow{8}{*}{AR} & {\scriptsize LlamaGen-L~\cite{sun2024autoregressive}} & 343M & 3.80 & 248.3 & 0.83 & 0.51 \\
&  & {\scriptsize LlamaGen-XL~\cite{sun2024autoregressive}} & 775M & 3.39 & 227.1 & 0.81 & 0.54 \\
&  & {\scriptsize LlamaGen-XXL~\cite{sun2024autoregressive}} & 1.4B & 3.09 & 253.6 & 0.83 & 0.53 \\
&  & {\scriptsize LlamaGen-3B~\cite{sun2024autoregressive}} & 3.1B & 3.05 & 222.3 & 0.80 & 0.58 \\

 &  & {\scriptsize LlamaGen-L-384~\cite{sun2024autoregressive}} & 343M & 3.07 & 256.1 & 0.83 & 0.52 \\
&  & {\scriptsize LlamaGen-XL-384~\cite{sun2024autoregressive}} & 775M & 2.62 & 244.1 & 0.80 & 0.57 \\
&  & {\scriptsize LlamaGen-XXL-384~\cite{sun2024autoregressive}} & 1.4B & 2.34 & 253.9 & 0.80 & 0.59 \\
&  & {\scriptsize LlamaGen-3B-384~\cite{sun2024autoregressive}} & 3.1B & 2.18 & 263.3 & 0.81 & 0.58 \\

\hline
\multirow{4}{*}{VQ~\cite{chang2022maskgit}} & \multirow{4}{*}{AR} & {\scriptsize RAR-B (ours)} & 261M & 1.95 & 290.5 & 0.82 & 0.58  \\
 & & {\scriptsize RAR-L (ours)} & 461M & 1.70 & 299.5 & 0.81 & 0.60  \\
 & & {\scriptsize RAR-XL (ours)} & 955M & 1.50 & 306.9 & 0.80 & 0.62 \\
 & & {\scriptsize RAR-XXL (ours)} & 1.5B & \textbf{1.48} & 326.0 & 0.80 & 0.63 \\

\end{tabular}
\caption{\textbf{ImageNet-1K $256\times 256$ generation results evaluated with ADM~\cite{dhariwal2021diffusion}.} ``type'' refers to the type of the generative model, where ``Diff.'' and ``Mask.'' stand for diffusion models and masked transformer models, respectively. ``VQ'' denotes discrete tokenizers and ``VAE'' stands for continuous tokenizers. ``-re'' stands for rejection sampling. ``-384'' denotes for generating images at resolution $384$ and resize back to $256$ for evaluation, as is used in~\cite{sun2024autoregressive}.
\vspace{-4ex}
}
\label{tab:imagenet_256}
\end{table}
\begin{table}
\centering
\tablestyle{3.0pt}{1.05}
\begin{tabular}{c|ccccc}
method & type & \#params & FID$\downarrow$ & steps & images/sec \\
\shline
DiT-XL/2~\cite{peebles2023scalable} & Diff. & 675M & 2.27 & 250 & 0.6 \\
TiTok-S-128~\cite{yu2024image} & Mask. & 287M & 1.97 & 64 & 7.8 \\
VAR-d30~\cite{tian2024visual} & VAR & 2.0B & 1.92 & 10 & 17.3 \\
MAR-B~\cite{li2024autoregressive} & MAR & 208M & 2.31 & 256 & 0.8 \\
RAR-B (ours) & AR & 261M & 1.95 & 256 & 17.0 \\
\hline
MAR-L~\cite{li2024autoregressive} & MAR & 479M & 1.78 & 256 & 0.5 \\
RAR-L (ours) & AR & 461M & 1.70 & 256 & 15.0 \\
\hline
MaskBit~\cite{weber2024maskbit} & Mask. & 305M & 1.52 & 256 & 0.7 \\
MAR-H~\cite{li2024autoregressive} & MAR & 943M & 1.55 & 256 & 0.3 \\
RAR-XL (ours) & AR & 955M & 1.50 & 256 & 8.3 \\
RAR-XXL (ours) & AR & 1.5B & 1.48 & 256 & 6.4
\end{tabular}
\caption{\textbf{Sampling throughput comparison (including de-tokenization process) categorized by methods with similar FID scores.} Throughputs are measured as samples generated per second on a single A100 using float32 precision and a batch size of $128$, based on their official codebases. For VAR~\cite{tian2024visual} and our RAR, KV-cache is applied. ``Diff.'' and ``Mask.'' refer to diffusion models and masked transformer models, respectively.
}
\vspace{-1ex}
\label{tab:sampling_speed}
\end{table}
\subsection{Main Results}
\label{sec:main_results}
We report RAR results against state-of-the-art image generators on ImageNet-1K $256\times 256$ benchmark~\cite{deng2009imagenet}.

As shown in~\tabref{tab:imagenet_256}, RAR achieves significantly better performance compared to previous AR image generators. Specifically, the most compact RAR-B with $261$M parameters only, achieves an FID score $1.95$, already significantly outperforming current state-of-the-art AR image generators LlamaGen-3B-384 ($3.1$B, FID $2.18$, crop size 384)~\cite{sun2024autoregressive} and Open-MAGVIT2-XL ($1.5$B, FID $2.33$)~\cite{luo2024open}, while using $91\%$ and $81\%$ fewer model parameters respectively. It also surpasses the widely used diffusion models such as DiT-XL/2 (FID $1.95$ \vs $2.27$) and SiT-XL (FID $1.95$ \vs $2.06$) while only using $39\%$ model parameters compared to them. 

In~\tabref{tab:imagenet_256}, we further explore RAR at different model sizes (from $261$M to $1.5$B), where we observe strong scalability behavior with consistent performance improvement as model size scales up.
Notably, the largest variant RAR-XXL sets a new state-of-the-art result on ImageNet benchmark, with an FID score $1.48$. When compared to the other two recent methods VAR~\cite{tian2024visual} and MAR~\cite{li2024autoregressive}, both of which attempt to amend AR formulation for better visual generation quality, RAR not only demonstrates a superior performance (FID $1.48$ from RAR \vs $1.73$ from VAR and $1.55$ from MAR), but also keeps the whole framework compatible with language modeling and thus is more friendly for adapting the mature optimization and speed-up techniques for large language models to visual generation~\cite{sun2024autoregressive}. 

\begin{figure*}[t!]
    \centering
    \begin{subfigure}[t!]{0.33\textwidth}
        \includegraphics[width=1.0\linewidth]{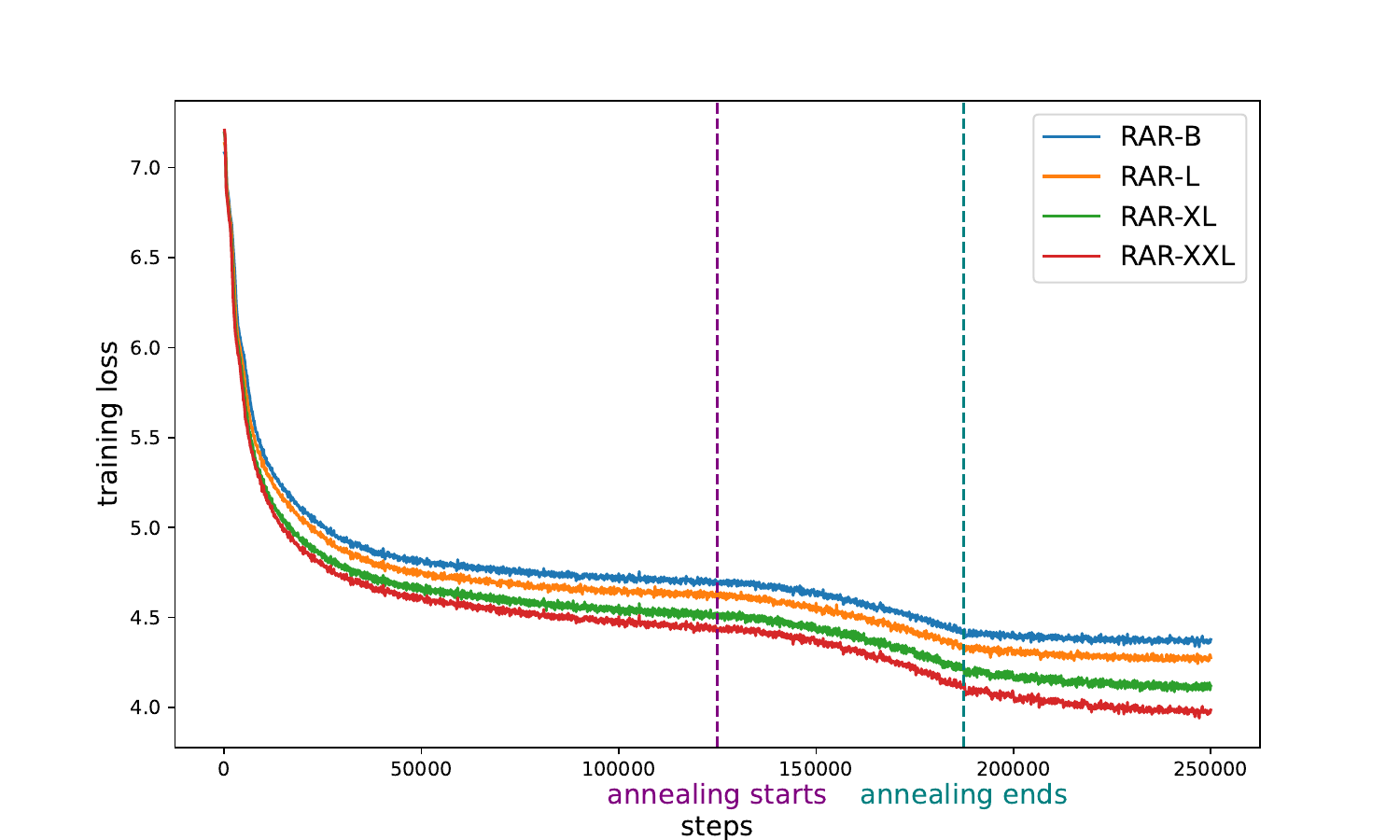}
        \caption{
        training losses
        }
        \label{fig:loss_curve}
    \end{subfigure}
    \begin{subfigure}[t!]{0.33\textwidth}
        \includegraphics[width=1.0\linewidth]{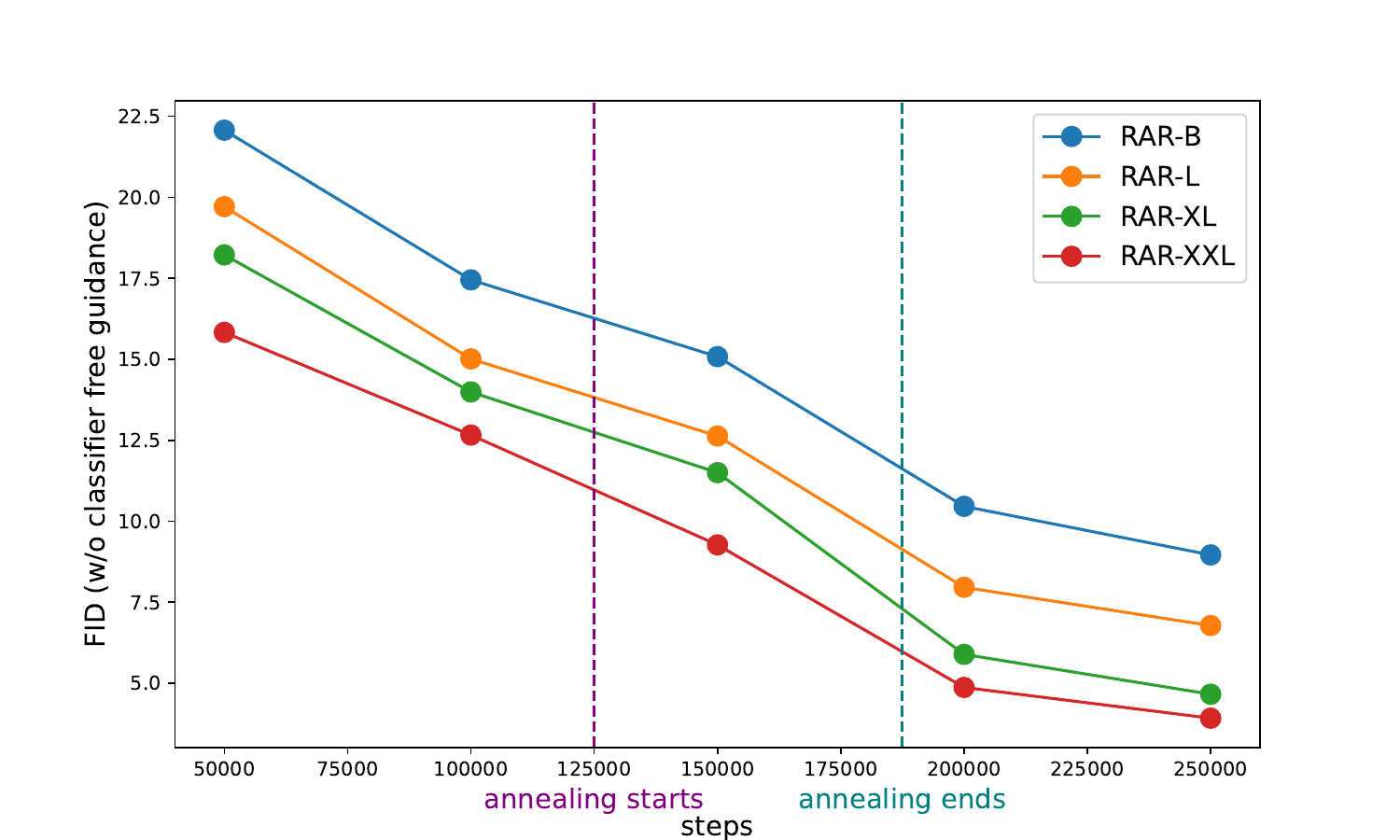}
        \caption{
        FID scores w/o classifier-free guidance 
        }
        \label{fig:fid_nocfg}
    \end{subfigure}
    \begin{subfigure}[t!]{0.33\textwidth}
        \includegraphics[width=1.0\linewidth]{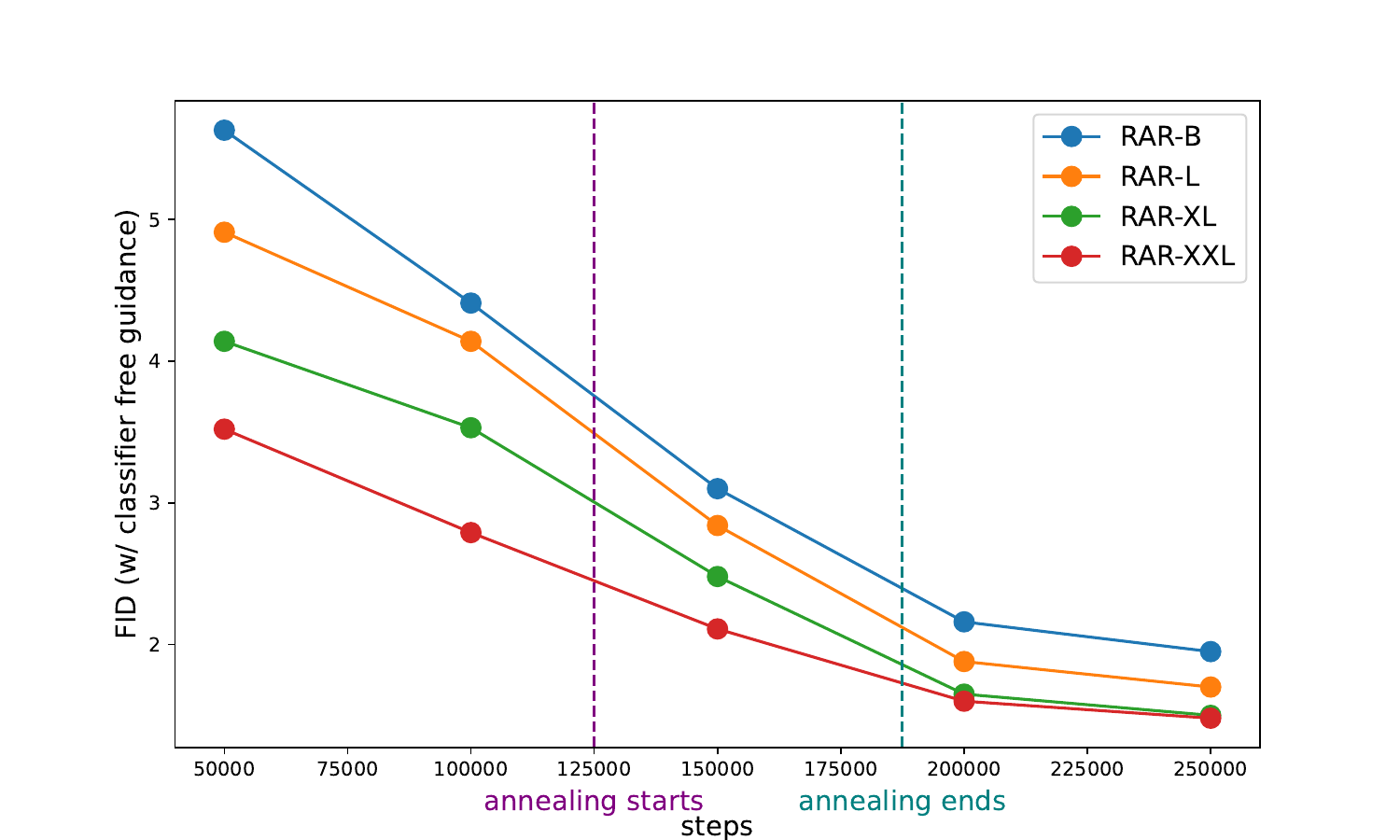}
        \caption{
        FID scores w/ classifier-free guidance 
        }
        \label{fig:fid_cfg}
    \end{subfigure}
    \caption{
    \textbf{Scaling behavior of RAR models.}
    The scaled-up RAR models demonstrate (a) reduced training losses, and improved FID scores both (b) without and (c) with classifier-free guidance.
    }
    \label{fig:scaling_curve}
\end{figure*}

Moreover, RAR demonstrates superior performance to state-of-the-art visual generators in different frameworks. It performs better against the leading autoregressive models, diffusion models and masked transformer models, surpassing LlamaGen-3B-384~\cite{sun2024autoregressive}, MDTv2-XL/2~\cite{gao2023masked} and MaskBit~\cite{weber2024maskbit} respectively (FID $1.48$ from RAR \vs $2.18$ from LlamaGen, $1.58$ from MDTv2, and $1.52$ from MaskBit). To the best of our knowledge, this is the first time that the language modeling style autoregressive visual generators outperform state-of-the-art diffusion models and masked transformer models.

\vspace{0.5ex}
\noindent \textbf{Sampling Speed.}
One key advantage of AR methods is their ability to leverage established optimization techniques from LLMs, such as KV-caching.
In~\tabref{tab:sampling_speed}, we compare the sampling speed (measured as images/sec) of RAR against other types of generative models, such diffusion models~\cite{peebles2023scalable}, masked transformers~\cite{yu2024image,weber2024maskbit}, VAR~\cite{tian2024visual}, and MAR~\cite{li2024autoregressive}. Among them, AR models (RAR) and VAR models (VAR-d30) are compatible with the KV-cache optimization, providing a significant advantage in generation speed over other methods.
As shown in~\tabref{tab:sampling_speed}, RAR achieves a state-of-the-art FID score while also significantly surpassing other  methods in generation speed. For instance, at an FID score around 1.5, MaskBit~\cite{weber2024maskbit} and MAR-H~\cite{li2024autoregressive} generate image samples at 0.7 and 0.3 images per second, respectively. In comparison, RAR-XL not only achieves a better FID score but can generate 8.3 high-quality visual samples per second—11.9$\times$ faster than MaskBit and 27.7$\times$ faster than MAR-H. The largest RAR variant, RAR-XXL, further improves the FID score while maintaining a notable speed advantage, being 9.1$\times$ faster than MaskBit and 21.3$\times$ faster than MAR-H. Additionally, RAR may benefit further from LLM optimization techniques such as vLLM~\cite{kwon2023efficient}, as seen with other AR methods~\cite{sun2024autoregressive}.

\vspace{0.5ex}
\noindent \textbf{Scaling Behavior.}
We study the scaling behavior of RAR. Specifically, we plot the training loss curves and FID score curves (with and without classifier-free guidance~\cite{ho2022classifier}) in~\figref{fig:scaling_curve}. As shown in the figure, we observe that RAR scales well at different model sizes, where larger model size leads to a consistently lower training loss and better FID score, regardless of using the enhancement of classifier-free guidance or not. We note that as RAR keeps the AR formulation and framework intact, it also inherits the scalability from AR methods.

\begin{figure}[t!]
    \centering
    \vspace{-6ex}
    \includegraphics[width=1.0\linewidth]{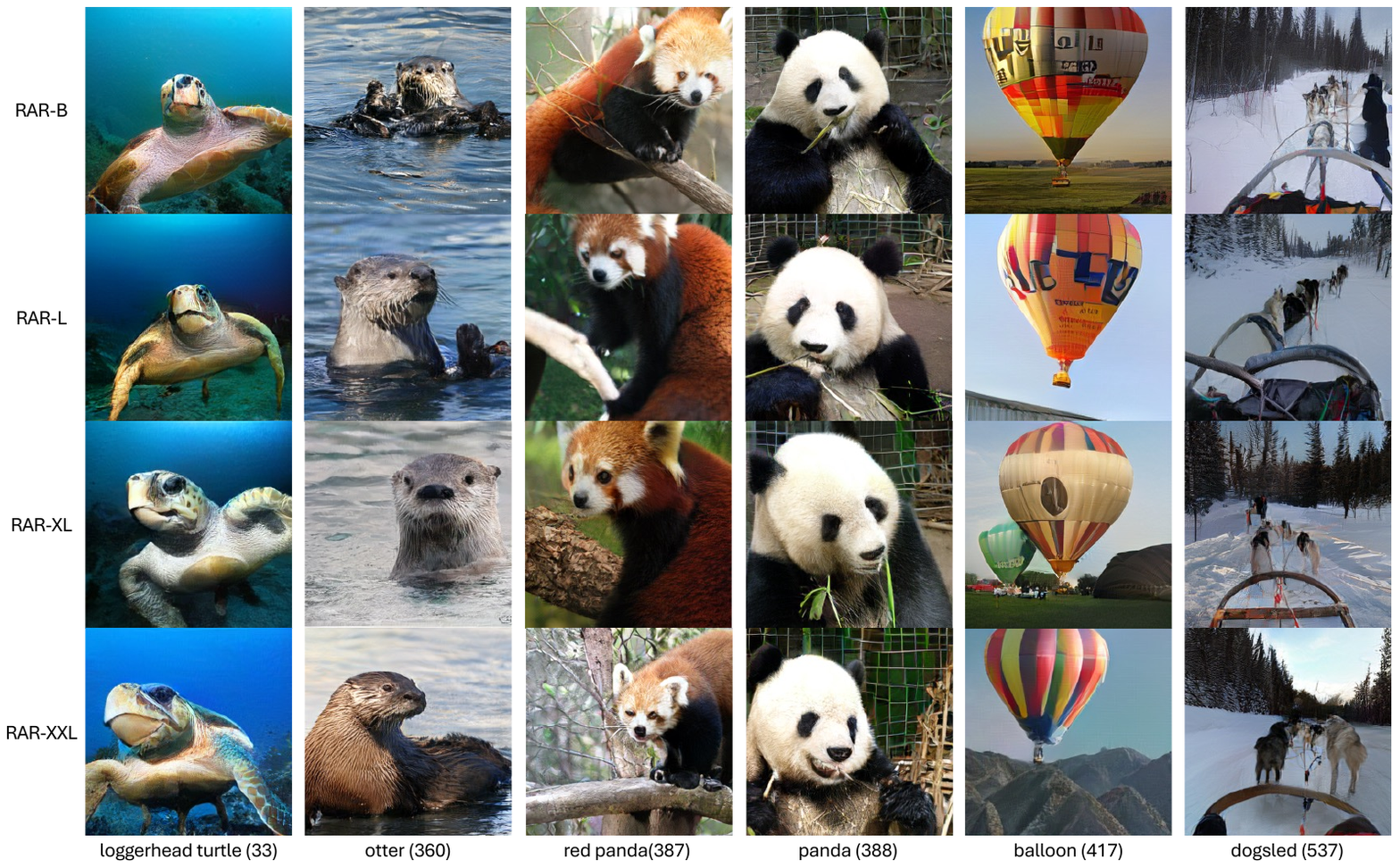}
    \vspace{-7ex}
    \caption{\textbf{Visualization of samples generated by RAR across various model sizes.}
    RAR generates high-quality visual samples across all model sizes. As model size increases, fidelity and diversity improve, especially in challenging classes (\eg, dogsled).
    }
    \label{fig:scaling_curve_vis}
\end{figure}

\vspace{0.5ex}
\noindent \textbf{Visualization.}
We visualize generated samples by different RAR variants in~\figref{fig:scaling_curve_vis}, which shows that RAR is capable of generating high-quality samples with great fidelity and diversity. More visualizations are provided in the appendix.

\section{Conclusion}

In this paper, we introduced a simple yet effective strategy to enhance the visual generation quality of language modeling-compatible autoregressive image generators. By employing a randomized permutation objective, our approach enables improved bidirectional context learning while preserving the autoregressive structure. Consequently, the proposed RAR model not only surpasses previous state-of-the-art autoregressive image generation models but also outperforms leading non-autoregressive transformer and diffusion models. We hope this research contributes to advancing autoregressive transformers toward a unified framework for visual understanding and generation.

\vspace{1ex}
\noindent \textbf{Acknowledgment.} We sincerely thank Tianhong Li for his insightful discussion and feedback on this project.
\clearpage
\setcounter{page}{1}
\maketitlesupplementary

\renewcommand{\thesection}{\Alph{section}}
\setcounter{section}{0}

\section*{Appendix}
\label{sec:appendix}

The supplementary material includes the following additional information:

\begin{itemize}
    \item \secref{sec:sup_hyper} provides the detailed hyper-parameters for the final RAR models.
    \item \secref{sec:sup_pseudo} provides the pseudo-code for randomized autoregressive modeling.
    \item \secref{sec:sup_vis_scan} visualizes the scan orders used in the ablation study.
    \item \secref{sec:sup_vis_samples} provides more visualization samples of RAR models.
\end{itemize}

\section{Hyper-parameters for Final RAR Models}
\label{sec:sup_hyper}

We list the detailed training hyper-parameters and sampling hyper-parameters for all RAR models in~\tabref{tab:hparams}.

\begin{table}[h!]
\centering

\tablestyle{5.0pt}{1.1}
\begin{tabular}{l|c}
config \quad\quad\quad\quad\quad\quad\quad\quad & value \\
\shline
\multicolumn{2}{c}{\textit{training hyper-params}} \\
\hline
optimizer & AdamW~\cite{kingma2014adam,loshchilov2017decoupled} \\
learning rate & 4e-4 \\
weight decay & 0.03 \\
optimizer momentum & (0.9, 0.96) \\
batch size & 2048 \\
learning rate schedule & cosine decay \\
ending learning rate & 1e-5 \\
total epochs & 400 \\
warmup epochs & 100 \\
annealing start epoch & 200 \\
annealing end epoch & 300 \\
precision & bfloat16 \\
max grad norm & 1.0 \\
dropout rate & 0.1 \\
attn dropout rate & 0.1 \\
class label dropout rate & 0.1 \\
\hline
\multicolumn{2}{c}{\textit{sampling hyper-params}}  \\
\hline
guidance schedule & pow-cosine~\cite{gao2023masked} \\
temperature & 1.0 (B) / 1.02 (L, XL, XXL) \\
scale power & 2.75 (B) / 2.5 (L) / 1.5 (XL) / 1.2 (XXL) \\
guidance scale & 16.0 (B) / 15.5 (L) / 6.9 (XL) / 8.0 (XXL) \\

\hline
\end{tabular}
\caption{\textbf{Detailed hyper-parameters for final RAR models.}
}
\label{tab:hparams}
\end{table}

\section{Pseudo-Code for RAR}
We provide a simple pseudo-code of RAR in PyTorch style in Algorithm~\ref{algo:pseduo_code}.
\label{sec:sup_pseudo}
\definecolor{commentcolor}{rgb}{0.2, 0.6, 0.2}
\definecolor{classcolor}{rgb}{0.7, 0.1, 0.2}
\definecolor{functioncolor}{rgb}{0.1, 0.1, 0.8}
\definecolor{keywordcolor}{rgb}{0.6, 0.2, 0.6}

\newcommand{\comment}[1]{\textcolor{commentcolor}{#1}}
\newcommand{\class}[1]{\textcolor{classcolor}{\textbf{#1}}}
\newcommand{\function}[1]{\textcolor{functioncolor}{\textbf{#1}}}
\newcommand{\keyword}[1]{\textcolor{keywordcolor}{\textbf{#1}}}

\begin{algorithm}[t]
\caption{PyTorch Pseudo-Code for Randomized AutoRegressive (RAR) Modeling
}
\label{algo:pseduo_code}
\begin{algorithmic}[0]
\footnotesize
\STATE \class{class} RAR(\keyword{nn.Module}): 
\STATE \quad \function{def} sample\_orders(self, tokens, global\_step): 
\STATE \quad \quad \comment{\# sample permutation order at training step global\_step.} 
\STATE \quad \quad orders = []
\STATE \quad \quad \comment{\# compute the randomized probability $r$ as in Eq.~\eqref{eq:r_schedule}.}
\STATE \quad \quad prob = 1.0 - min(1.0, max(0.0, (global\_step - self.anneal\_start) / (self.anneal\_end - self.anneal\_start)))
\STATE \quad \quad \keyword{for} b \keyword{in} \keyword{range}(tokens.shape[0]):
\STATE \quad \quad \quad \keyword{if} random.random() \textless  prob:
\STATE \quad \quad \quad \quad \comment{\# random permutation.}
\STATE \quad \quad \quad \quad orders.append(torch.randperm(tokens.shape[1]))
\STATE \quad \quad \quad \keyword{else}:
\STATE \quad \quad \quad \quad \comment{\# raster order (no permutation).}
\STATE \quad \quad \quad \quad orders.append(torch.arange(tokens.shape[1]))
\STATE \quad \quad \keyword{return} torch.stack(orders)
\STATE 
\STATE \quad \function{def} permute(self, inputs, orders):
\STATE \quad \quad \comment{\# permute inputs based on orders.}
\STATE \quad \quad B, L = inputs.shape[:2]
\STATE \quad \quad indices = torch.arange(B).unsqueeze(1).expand(-1, L)
\STATE \quad \quad \keyword{return} x[indices, orders]
\STATE 
\STATE \quad \function{def} forward(self, tokens, condition, global\_step):
\STATE \quad \quad \comment{\# get permutation orders.}
\STATE \quad \quad orders = self.sample\_orders(global\_step, tokens)
\STATE \quad \quad \comment{\# permute labels for next-token prediction.}
\STATE \quad \quad labels = self.permute(tokens.clone(), orders)
\STATE \quad \quad \comment{\# token embeddings with positional embedding.}
\STATE \quad \quad x = self.tok\_emb(tokens) + self.pos\_emb
\STATE \quad \quad \comment{\# permute the token orders.}
\STATE \quad \quad x = self.permute(x, orders)
\STATE \quad \quad \comment{\# add target-aware postional embedding as in Eq.~\eqref{eq:target_aware}.}
\STATE \quad \quad target\_pos\_emb = self.target\_pos\_emb.repeat(x.shape[0], 1, 1)
\STATE \quad \quad target\_pos\_emb = self.permute(target\_pos\_emb, orders)
\STATE \quad \quad \comment{\# shifting so each token will see next-token's embedding.}
\STATE \quad \quad target\_pos\_emb = target\_pos\_emb[:, 1:]
\STATE \quad \quad x = torch.cat([x[:, :-1] + target\_pos\_emb, x[:, -1:]], dim=1)
\STATE \quad \quad \comment{\# transformer forwarding.}
\STATE \quad \quad pred = self.transformers(x, condition)
\STATE \quad \quad \comment{\# next token prediction loss.}
\STATE \quad \quad loss = nn.CrossEntropy(pred[:, :-1], labels[:, 1:])
\STATE \quad \quad \keyword{return} loss
\end{algorithmic}
\end{algorithm}

\section{Visualization of Scan Orders}
\label{sec:sup_vis_scan}
\begin{figure*}[t!]
    \centering
    \vspace{-3ex}
    \begin{subfigure}[b]{0.25\textwidth}
        \includegraphics[width=\textwidth]{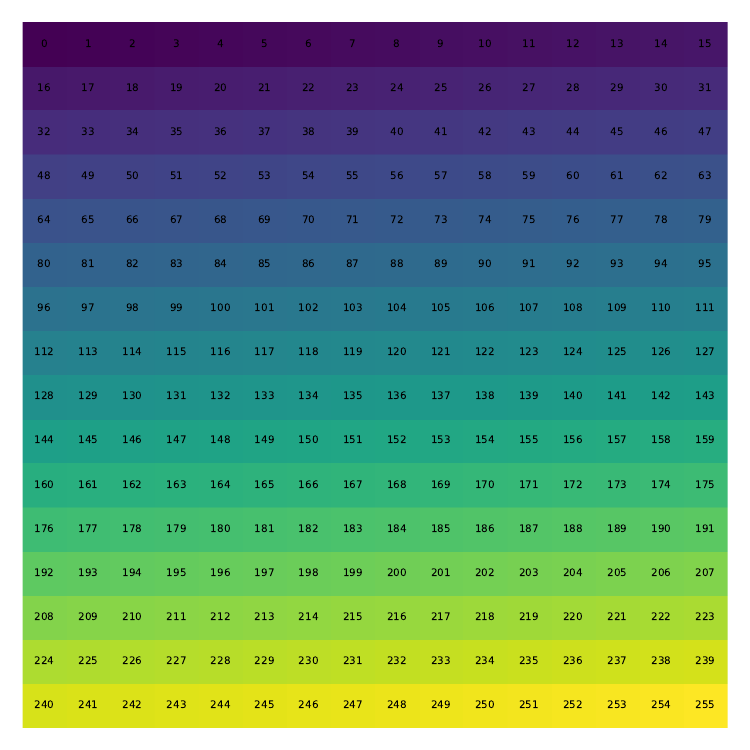}
        \caption{row-major}
        \label{fig:pdf1}
    \end{subfigure}
    \hfill
    \begin{subfigure}[b]{0.25\textwidth}
        \includegraphics[width=\textwidth]{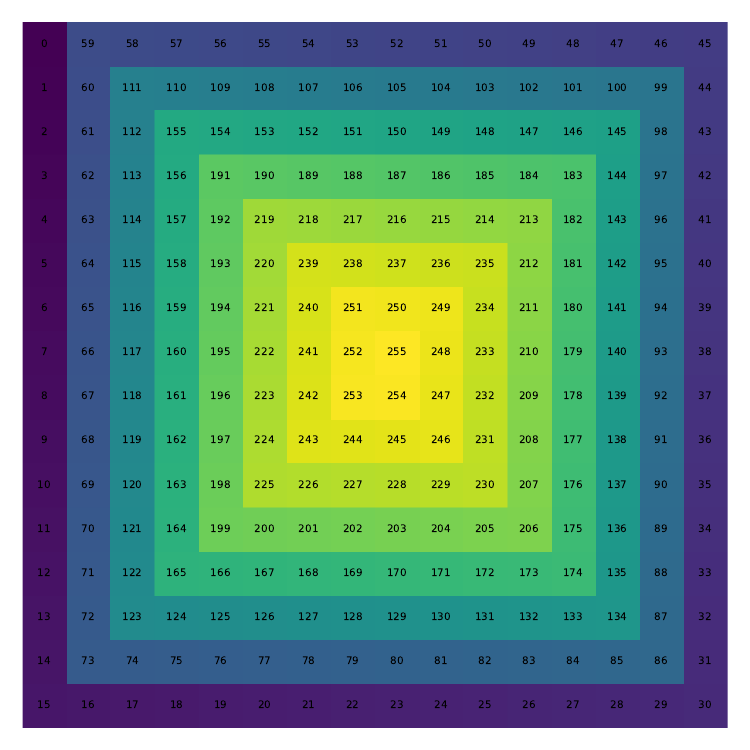}
        \caption{spiral in}
        \label{fig:pdf2}
    \end{subfigure}
    \hfill
    \begin{subfigure}[b]{0.25\textwidth}
        \includegraphics[width=\textwidth]{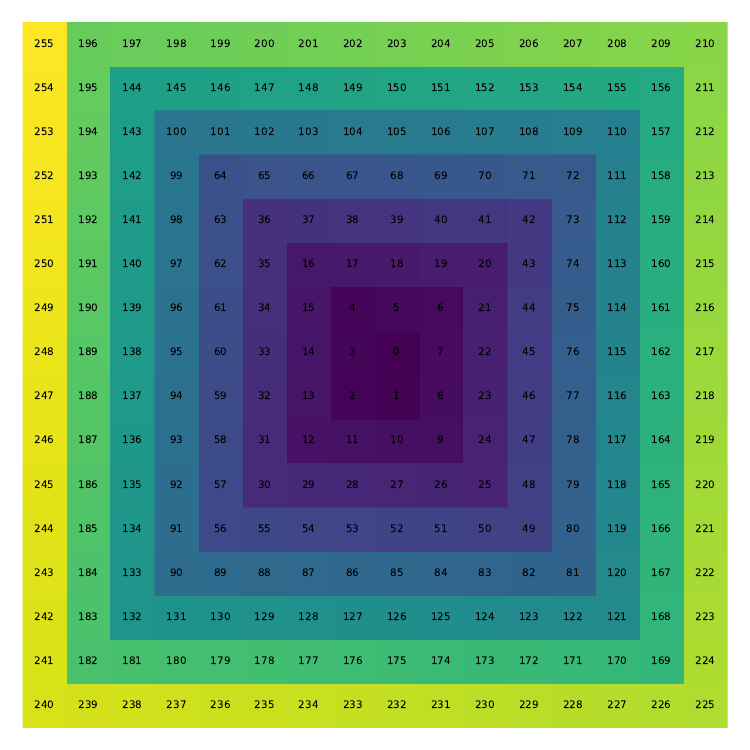}
        \caption{spiral out}
        \label{fig:pdf3}
    \end{subfigure}


    \begin{subfigure}[b]{0.25\textwidth}
        \includegraphics[width=\textwidth]{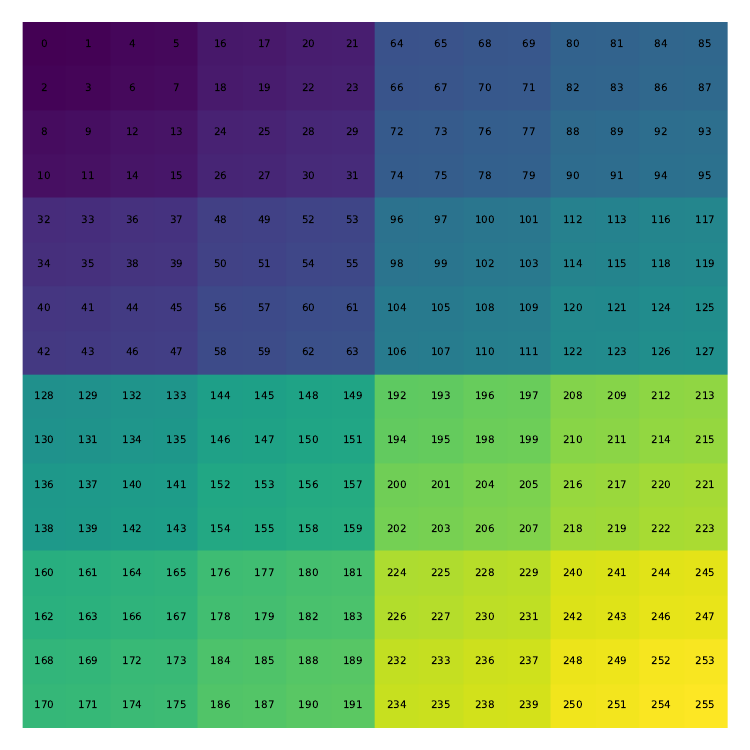}
        \caption{z-curve}
        \label{fig:pdf4}
    \end{subfigure}
    \hfill
    \begin{subfigure}[b]{0.25\textwidth}
        \includegraphics[width=\textwidth]{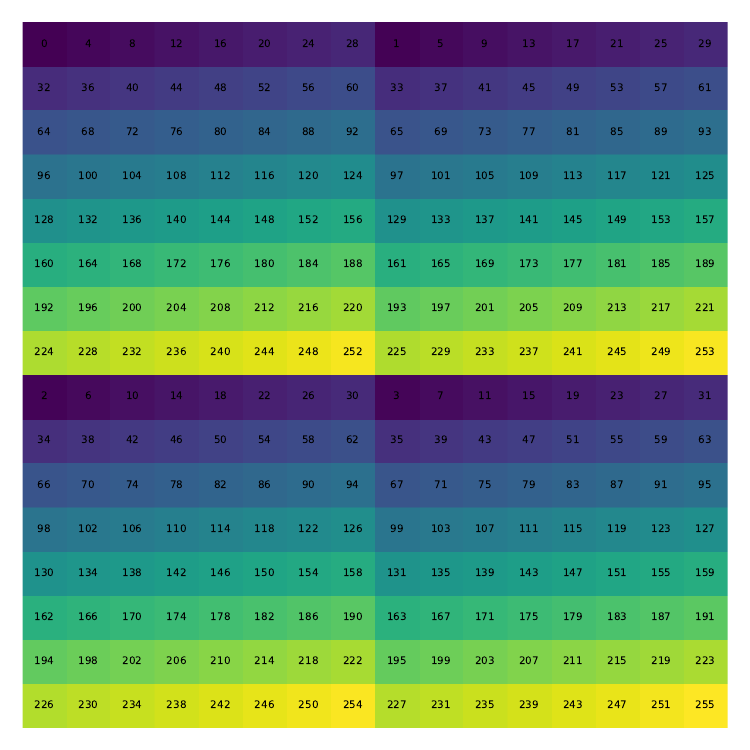}
        \caption{subsample}
        \label{fig:pdf5}
    \end{subfigure}
    \hfill
    \begin{subfigure}[b]{0.25\textwidth}
        \includegraphics[width=\textwidth]{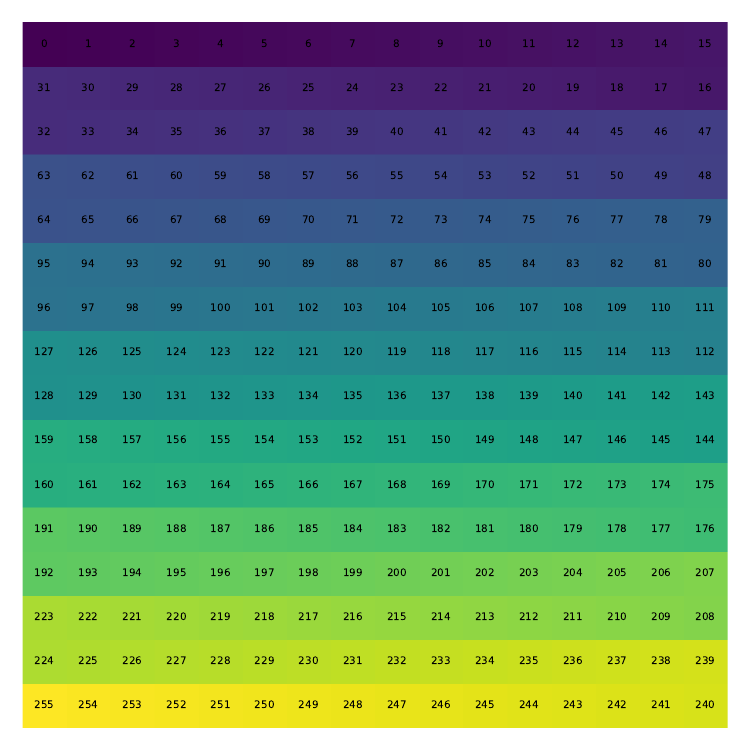}
        \caption{alternate}
        \label{fig:pdf6}
    \end{subfigure}

    \caption{\textbf{Different scan orders for a $16\times 16$ grid (256 tokens).} The number indicates the token's indices in the scanning order.}
    \label{fig:all_scan_orders}
\end{figure*}

We visualize the 6 scan orders studied in the main paper (~\tabref{tab:scan_orders}) in~\figref{fig:all_scan_orders}.

\section{Visualization on Generated Samples}
\label{sec:sup_vis_samples}
\begin{figure*}[t!]
    \centering
    \vspace{-8ex}
    \includegraphics[width=0.9\linewidth]{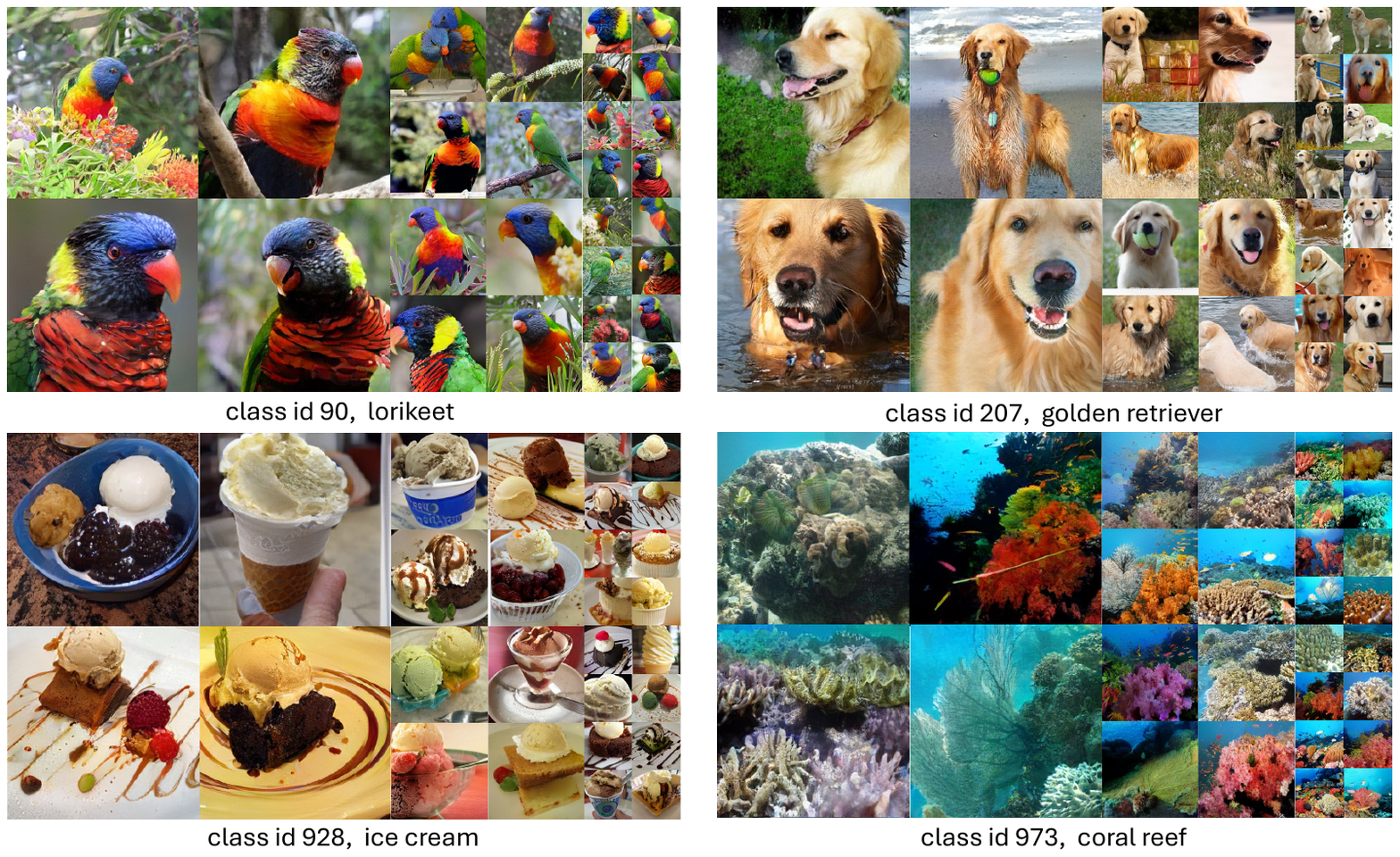}
    \vspace{-8ex}
    \caption{
    \textbf{Visualization samples from RAR.} RAR is capable of generating high-fidelity image samples with great diversity.
    }
    \label{fig:vis_more}
\end{figure*}

\begin{figure*}[t!]
    \centering
    \vspace{-8ex}
    \includegraphics[width=0.9\linewidth]{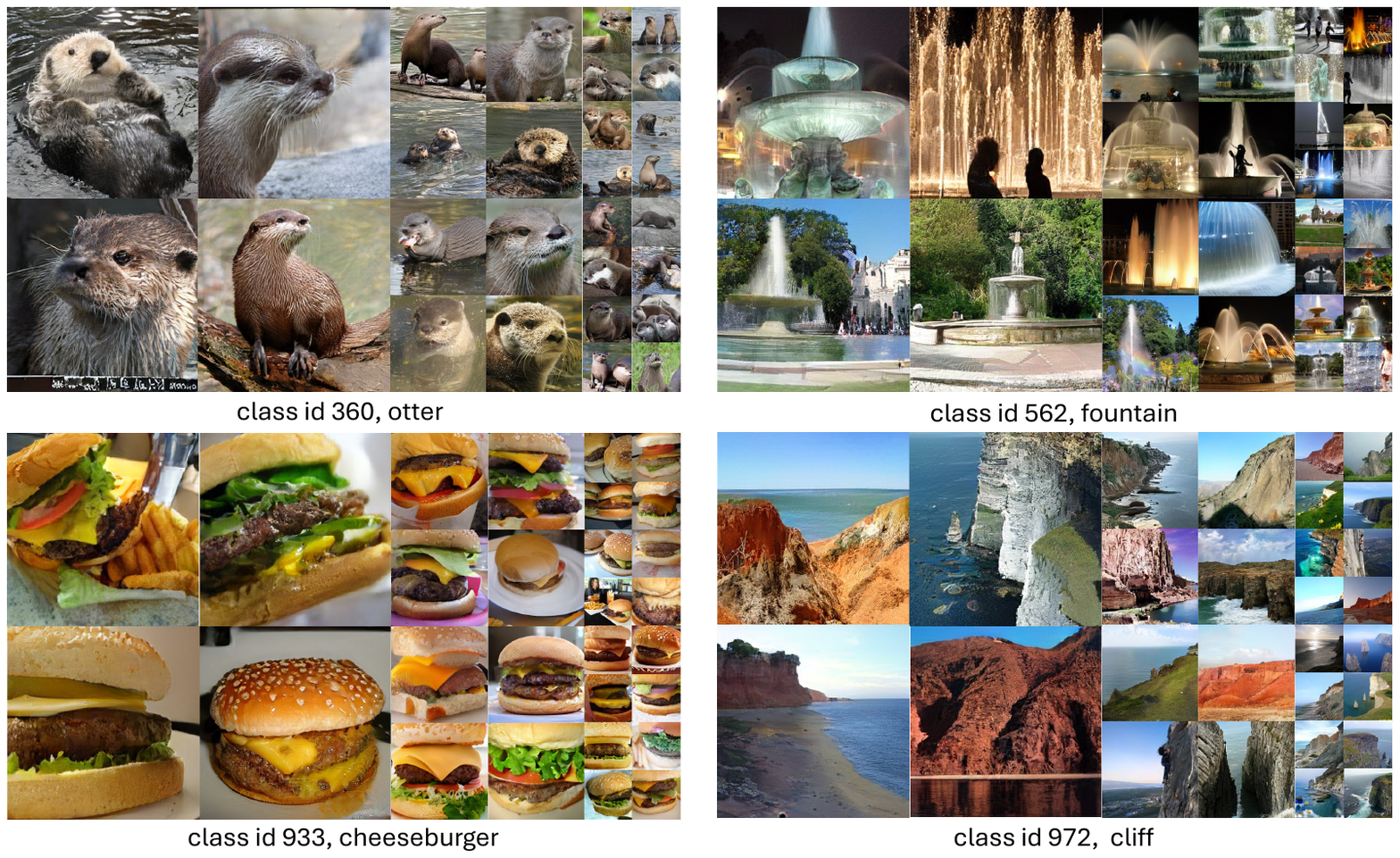}
    \vspace{-8ex}
    \caption{
    \textbf{Visualization samples from RAR.} RAR is capable of generating high-fidelity image samples with great diversity.
    }
    \label{fig:vis_more2}
\end{figure*}

\begin{figure*}[t!]
    \centering
    \vspace{-10ex}
    \includegraphics[width=0.9\linewidth]{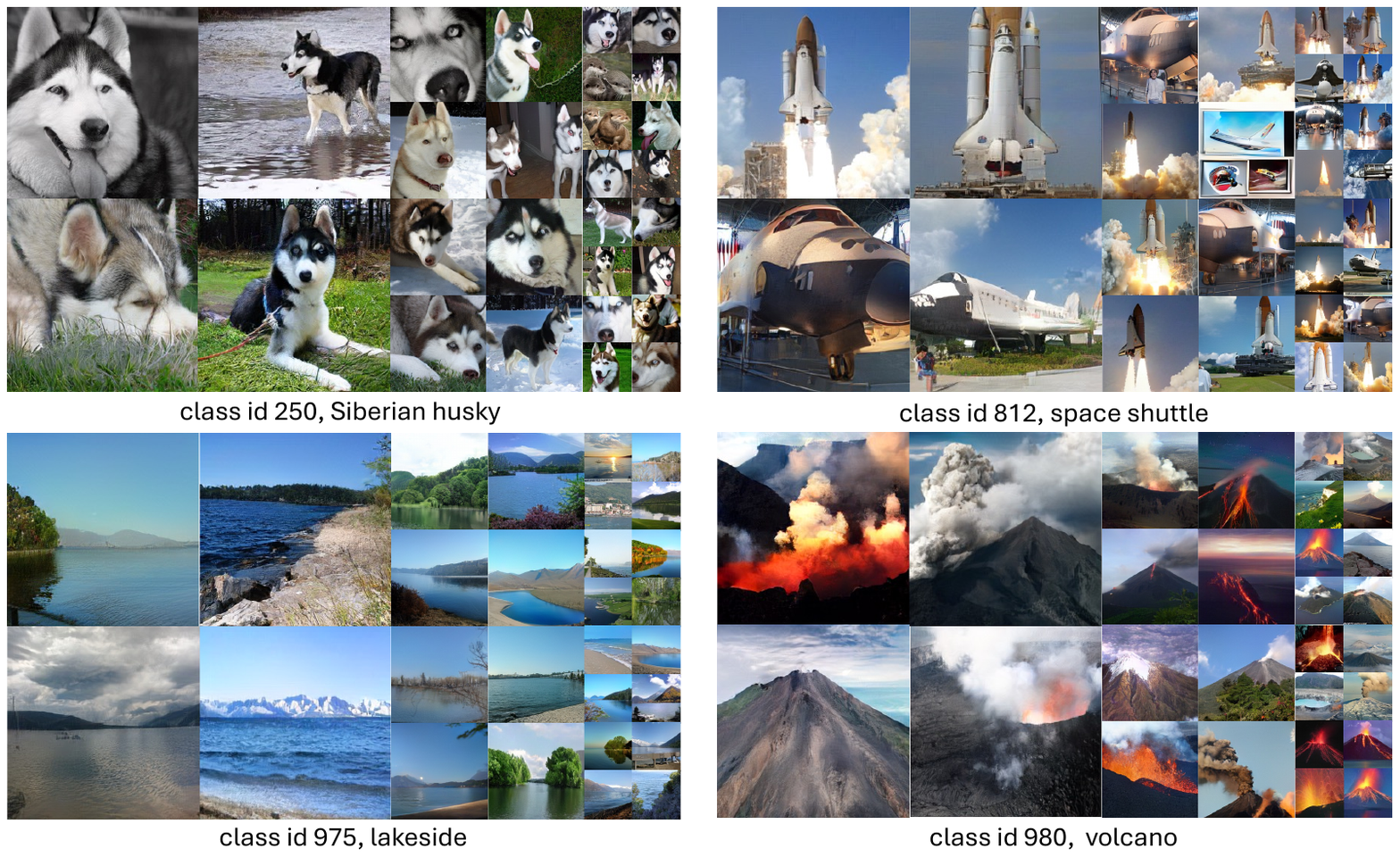}
    \vspace{-8ex}
    \caption{
    \textbf{Visualization samples from RAR.} RAR is capable of generating high-fidelity image samples with great diversity.
    }
    \label{fig:vis_more3}
\end{figure*}

We provide visualization results in~\figref{fig:vis_more},~\figref{fig:vis_more2}, and~\figref{fig:vis_more3}.

\clearpage
\clearpage

{
    \small
    \bibliographystyle{ieeenat_fullname}
    \bibliography{main}
}


\end{document}